\documentclass{article}
\usepackage{spconf,amsmath,graphicx,hyperref}
\usepackage{amsmath}    % 基础数学符号
\usepackage{amssymb}    % 提供 \mathbb 等符号
\usepackage{bm}         % 加粗数学符号（可选）
\usepackage{multirow} % 用于多行/多列合并
\usepackage{booktabs} % 美化表格线
\usepackage{array} % 对齐控制
\usepackage{graphicx} % 用于调整表格大小
\usepackage{makecell}
\usepackage{bm} % 粗体数学符号
\usepackage{xcolor} % 必须的
\usepackage{soul}   % 提供高亮功能
\usepackage[export]{adjustbox} % 关键宏包
\usepackage{tikz} % 用于绘制色块
\usepackage[utf8]{inputenc}
\usepackage{anyfontsize}
\definecolor{highlighter_yellow}{RGB}{255,255,200}  % 柔和的淡黄色
\definecolor{highlighter_red}{RGB}{255,200,200}     % 更柔和的粉红色
\definecolor{highlighter_orange}{RGB}{255,210,150}  % 更柔和的浅橙色
\newcommand{\yellowhl}[1]{\sethlcolor{highlighter_yellow}\hl{#1}}
\newcommand{\redhl}[1]{\sethlcolor{highlighter_red}\hl{#1}}
\newcommand{\orangehl}[1]{\sethlcolor{highlighter_orange}\hl{#1}}

\newcommand{\showyellow}{\textcolor{highlighter_yellow}{\rule{0.4cm}{0.2cm}}}
\newcommand{\showred}{\textcolor{highlighter_red}{\rule{0.4cm}{0.2cm}}}
\newcommand{\showorange}{\textcolor{highlighter_orange}{\rule{0.4cm}{0.2cm}}}

\title{Gen3d:Generating Domain-free 3D Scenes from a Single Image} 

\name{
    \begin{tabular}{c} 
    Yuxin Zhang\textsuperscript{1,2},
    Ziyu Lu\textsuperscript{1,2},
    Hongbo Duan\textsuperscript{1,2},
    Keyu Fan\textsuperscript{1} \\
    Pengting Luo\textsuperscript{2},
    Peiyu Zhuang\textsuperscript{2},
    Mengyu Yang\textsuperscript{1,2},
    Houde Liu\textsuperscript{1,*}\thanks{ *Corresponding author: \texttt{liu.hd@sz.tsinghua.edu.cn}}
    \end{tabular}
}

\address{
    \textsuperscript{1} Shenzhen International Graduate School, Tsinghua University, Shenzhen, China \\ % 机构1
    \textsuperscript{2} Central Media Technology Institute, Huawei, Shenzhen, China \\ % 机构2
}
\ninept

\begin{document}

\ninept

\maketitle

\begin{abstract}
Despite recent advancements in neural 3D reconstruction, the dependence on dense multi-view captures restricts their
broader applicability.Additionally, 3D scene generation is vital for advancing embodied AI and world models, which depend on diverse, high-quality scenes for learning and evaluation.In this work, we propose Gen3d, a novel method for generation of high-quality, wide-scope, and generic 3D scenes
from a single image.
After the initial point cloud is created by lifting the RGBD image, Gen3d maintains and expands its world model.The 3D scene is finalized through optimizing a Gaussian splatting representation.
Extensive experiments on diverse datasets demonstrate the strong generalization capability and superior performance of our method in generating a world model and 
Synthesizing high-fidelity and consistent novel views. 
% Our project webpage and code are available at https://github.com/zhangyuxin-1/gen3d.
\end{abstract}
\section{Introduction}
\label{sec:intro}

With the advent of commercial mixed reality platforms and the rapid innovations in 3D graphics technology, high-quality 3D scene generation has become one of the most important problem in computer vision for applications such as immersive media, robotics, autonomous driving, and embodied AI.Previous approaches mainly include Geometric reconstruction and video generation methods.Geometric reconstruction methods suffer from severe artifacts at viewpoints unseen in input views, while video generation methods lack good geometric consistency in their outputs.

In this work, we propose a pipeline called Gen3d.To address the artifact generation issue of geometric reconstruction methods, Gen3d adopts a hierarchical strategy and leverages Stable Diffusion \cite{1} and 3D Gaussian splatting \cite{2} to generate a wide variety of high-quality 3D scenes with geometric consistency from multiple types of inputs, such as text, RGB images, and RGBD images. 
% that utilizes Stable Diffusion \cite{1} and 3D Gaussian splatting \cite{2} to create diverse high-quality 3D scenes from various types of inputs such as text, RGB, and RGBD.

Gen3d first decomposes the input image into foreground and background layers via depth maps (input or from depth models).The background is inpainted via Lama-ControlNet inpainting\cite{26} guided image inpainting, and initial point clouds are generated for both layers. 
The camera then moves along a predefined trajectory. At each pose, visible parts of the point cloud are projected onto the new view, and a Stable Diffusion-based inpainting network synthesizes a complete image from this partial projection. The result, along with its depth map, is lifted back to 3D space, with point positions refined via an alignment algorithm. 
Finally, the aggregated point cloud initializes a 3D Gaussian splatting representation, which resolves depth inconsistencies and enables high-quality novel view synthesis beyond traditional point-based methods. 

In summary, our contributions are as follows.

• \textbf{We introduce Gen3d, a domain-free,high-quality 3D scene generation method,} achieving better domain generalization in 3D scene generation by leveraging the power of Stable Diffusion, depth estimation, and explicit 3D representation.

• To generate multi-view images, \textbf{we adopt a hierarchical strategy} that first constructs a point cloud as a geometric prior for each novel view to generate novel view images. The generated images are then seamlessly integrated to form a coherent 3D scene.

• Our model provides users with the ability to \textbf{create 3D scenes in various ways by supporting different input types,} such as text, RGB, and RGBD.

\section{Raleted work}
\label{sec:format}
\textbf{Novel view generation.}Given a single image, early methods infer 3D scene representations and use them to render novel views. These representations include point clouds \cite{3,6}, multi-plane images \cite{7,8}, depth maps\cite{9}, and meshes \cite{10}. Despite enabling fast rendering, these representations limit camera movement due to their finite spatial extent. 
In addition, they only supported generating views within small viewpoint changes w.r.t. the input image, as they only built single static scene representations that do not go beyond the input image. Gen3d focus on a generative task to support creating many connected scenes rather than a single one.

\noindent\textbf{3D scene generation.}Recently, advancements in scene generation have predominantly centered on modeling isolated, local 3D environments. A significant number of these efforts explicitly target indoor settings \cite{11,20}, while another line of research concentrates predominantly on outdoor scenarios \cite{12}. This thematic specialization limits the generalizability of such approaches across diverse environmental contexts.

\noindent\textbf{Video generation.}Recent improvements in video generation \cite{13} have led to interest in whether these models can also be used as scene generators. Several works have attempted to add camera control, allowing a user to ``move'' through the scene \cite{14}. While these are promising, they currently do not guarantee 3D consistency.

\noindent\textbf{Fast 3D scene representations.}3D Gaussian Splatting(3DGS) excels in high-fidelity real-time rendering by explicitly modeling scenes as anisotropic 3D Gaussians—each defined by position, covariance, opacity, and spherical harmonics coefficients. Unlike implicit volumetric methods requiring costly ray marching, 3DGS enables efficient GPU-based rasterization, achieving real-time high-resolution view synthesis. Its adaptive optimization dynamically adjusts Gaussians to capture fine details. Leveraging these advantages (efficiency, explicit control, and rendering quality), we adopt 3DGS as our core representation for sparse-input 3D reconstruction.

\begin{figure*}[htb]
  \centering
  \includegraphics[width=\textwidth]{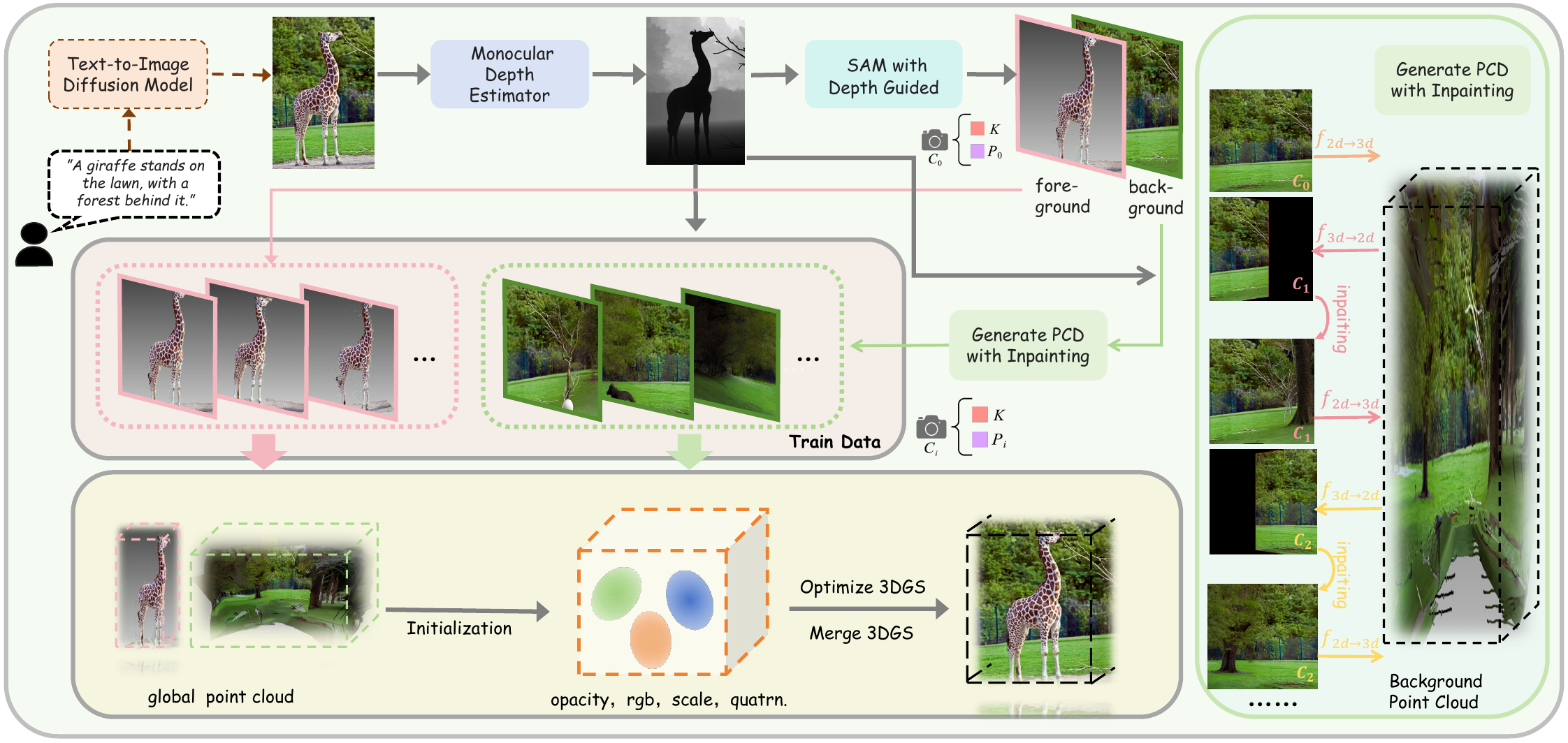} % 使用 \textwidth 确保图片宽度适合双栏
  \caption{\textbf{Gen3d pipeline.}Initially, the input 2D image is segmented into two distinct components: the foreground objects and the background.  
We adopt methodologies such as the Stable Diffusion model and monocular depth estimation to enhance point cloud coverage and facilitate the construction of larger-scale scenes.Subsequently, we employ the point cloud alongside the reprojected images to optimize a set of Gaussian splats, further refining the resulting 3D scene.}
  \label{fig:pipeline}
\end{figure*}

\section{Methodology}
\label{sec:majhead}

Given an input image or text prompt, our purpose is to generate realistic and high-quality 3D scenes that are conditioned on this input. In scenarios where only a text prompt is provided,  Gen3d  is capable of producing scenes that are semantically related to the given text. Furthermore, Gen3d excels in creating specified scenes based on text prompts while preserving the stylistic elements of the input image.

\vspace{-2pt} 
\subsection{Single-view Layer Generation}
\label{ssec:subhead1}

To decompose a scene into layered representations from a single view, we propose a hybrid approach combining depth-prior-guided segmentation with diffusion-based inpainting.As illustrated in Fig.~\ref{fig:pipeline}, our framework first generates foreground object masks $\mathcal{M}_{fg}$ and background masks $\mathcal{M}_{bg}$ using depth-aware Segment Anything Model (SAM)\cite{16}, then reconstructs occluded regions in the background layer through text-conditioned inpainting.

\noindent{\bfseries Depth-guided Mask Generation.}Given an input image $I \in  \mathbb{R}^{H\times W\times 3}$, we estimate its depth map $D \in \mathbb{R}^{H\times W}$ using Moge2\cite{15}. The SAM model produces $N$ candidate masks $\{\mathbf{M}_i\}_{i=1}^N$ with associated confidence scores $\{s_i\}_{i=1}^N$. We filter these masks through:
\begin{equation}    
\mathcal{V} = \{\mathbf{M}_i | s_i > \tau_{iou} \wedge \text{area}(\mathbf{M}_i) \in [A_{min}, A_{max}]\}
\end{equation}
where $\tau_{iou}=0.85$ and $A_{min/max}$ are set as 0.5\% / 60\% of image area. Each valid mask is then classified by depth consistency:
\begin{equation}
\mathbf{M}_{fg} = \bigcup_{\mathbf{M}_i \in \mathcal{V}} \left\{ \mathbf{M}_i \mid \text{median}(D|_{\mathbf{M}_i}) < \theta_d \right\}
\end{equation}
with $\theta_d$ dynamically set as the 35\% of $D$. The background mask is derived as $\mathbf{M}_{bg} = \mathbf{1} - \mathbf{M}_{fg}$.

\noindent\textbf{Occlusion-aware Inpainting.}
For regions $\Omega = \mathbf{M}_{fg} \cap \{D > \theta_d\}$, we employ Lama-ControlNet  with two key modifications:

(1)Depth-conditioned attention:The cross-attention layers in Stable Diffusion are modulated by normalized depth values $\tilde{D} = (D - D_{min})/(D_{max} - D_{min})$.

(2)Textual prompting: Background inpainting uses the template ``high-resolution [scene category] background with [dominant colors] colors, photorealistic, 8K'' where bracketed terms are auto-filled by CLIP\cite{17}.

\vspace{-2pt} 
\subsection{world generation}
To generate a multi-view consistent 3D point cloud, we first create the initial point cloud.Subsequently, while moving the camera,we aggregate both the new points and initial points by moving back and forth between the 3D space and the camera plane, with the reconstruction results situated in a unified world coordinate system.The overall process of point cloud construction is illustrated in Figure~\ref{fig:pipeline}.

\noindent\textbf{Initialization.} A point cloud generation starts from lifting the pixels of the initial image. If the user provides a text prompt as input, a latent diffusion model is utilized to generate an image relevant to the given text, and the metric depth map is estimated using  Moge2. We denote the generated or received RGB image and the corresponding depth map as $\mathbf{I}_0 \in \mathbb{R}^{3 \times H \times W}$ and $\mathbf{D}_0 \in \mathbb{R}^{H \times W}$, respectively, where $H$ and $W$ represent the height and the width of the image. The camera intrinsic matrix and the extrinsic matrix of $\mathbf{I}_0$ are denoted as $\mathbf{K}$ and $\mathbf{P}_0$, respectively. For the case where $\mathbf{I}_0$ and $\mathbf{D}_0$ are generated from the diffusion model, we set the values of $\mathbf{K}$ and $\mathbf{P}_0$ by convention according to the size of the image.

From the RGBD image $[\mathbf{I}_0, \mathbf{D}_0]$, we lift the pixels into the 3D space, where the lifted pixels form a point cloud. The generated initial point cloud using the first image is defined as $\mathcal{P}_0$:
\begin{equation}
\mathcal{P}_0 = f_{\text{2d}\rightarrow\text{3d}} \left( [\mathbf{I}_0, \mathbf{D}_0], \mathbf{K}, \mathbf{P}_0 \right),
\end{equation}
where $f_{\text{2d}\rightarrow\text{3d}}$ is the function that lifts pixels from the RGBD image $[\mathbf{I}, \mathbf{D}]$ to the point cloud.

\noindent\textbf{Point Cloud Augmentation and aggregation.} We sequentially attach points to the original point cloud to construct a large-scale 3D scene. Specifically, we define a  counterclockwise rotating camera trajectory trajectory of length $N$, where $\mathbf{P}_i$ denotes the camera position and pose at the $i$-th index. At each step, we inpaint and lift the missing pixels.We leverage the representational power of Stable Diffusion for the image inpainting task.
Specifically,At step $i$, we first move and rotate the camera from the previous position $\mathbf{P}_{i-1}$ to $\mathbf{P}_i$. The coordinate system is transformed from world coordinates to the current camera coordinates, followed by projection onto the camera plane using the intrinsic matrix $\mathbf{K}$ and the extrinsic matrix $\mathbf{P}_i$.
We denote the projected image at camera $\mathbf{P}_i$ as $\hat{\mathbf{I}}_i$. Due to the change in camera position and pose, certain regions in $\hat{\mathbf{I}}_i$ cannot be filled from the existing point cloud. We define a binary mask $\mathbf{M}_i$ to indicate the filled regions: $\mathbf{M}_i$ equals 1 if the corresponding pixel is filled by existing points, and 0 otherwise. The Stable Diffusion inpainting model ($S$) is employed to generate a realistic image $\mathbf{I}_i$ from the incomplete image $\hat{\mathbf{I}}_i$ and the mask $\mathbf{M}_i$. The corresponding depth map $\hat{\mathbf{D}}_i$ is estimated using Moge2.

Note that the monocular depth estimation model provides metric depth values. We then lift pixels to 3D space using the inpainted image $\mathbf{I}_i$ and its corresponding depth map $\mathbf{D}_i$. To save memory consumption and represent efficiently, only pixels in the inpainted regions ($\mathbf{M}_i = 0$) are lifted.

Compared to approaches that train generative models to simultaneously produce both RGB and depth maps,such as RGBD2\cite{18} employing off-the-shelf depth estimation methods yields more accurate and generalizable depth maps, as these models are trained on large and diverse datasets. However, since $\mathbf{D}_0, \mathbf{D}_1, \dots, \mathbf{D}_{i-1}$ are not considered when estimating $\mathbf{D}_i$, an inconsistency arises when integrating new points $\hat{\mathcal{P}}_i$. 

To address this issue, we adjust the points in $\hat{\mathcal{P}}_i$ within the 3D space to ensure smooth integration between the existing point cloud $\mathcal{P}_{i-1}$ and the new points $\hat{\mathcal{P}}_i$. Specifically, we extract the region where the mask boundary changes ($|\nabla \mathbf{M}_i| > 0$) to identify corresponding points in both $\mathcal{P}_{i-1}$ and $\hat{\mathcal{P}}_i$. We then compute a displacement vector from $\hat{\mathcal{P}}_i$ to $\mathcal{P}_{i-1}$. However, naively moving points may distort the geometry of the lifted point cloud and cause misalignment with the inpainted image. To mitigate this, we impose constraints on point movement and employ an interpolation algorithm to preserve the overall structure.

First, we constrain each point in $\hat{\mathcal{P}}_i$ to move along the ray originating from the camera center to its corresponding pixel. We locate the closest point in $\mathcal{P}_{i-1}$ along this ray and quantify the required depth adjustment. This constraint ensures that the visual content of the RGB image $\mathbf{I}_i$ remains consistent despite 3D point adjustments. Second, we assume depth values remain unchanged on the opposite side of the mask boundary. For points without ground-truth correspondences (i.e., where $\mathbf{M}_i = 0$), we compute depth changes via linear interpolation. Smooth interpolation alleviates artifacts caused by abrupt movements.

The aligned point cloud is combined with the original as follows:
\begin{equation}
\mathcal{P}_i = \mathcal{P}_{i-1} \cup \mathcal{W} \left( \hat{\mathcal{P}}_i \right),
\end{equation}
where $\mathcal{W}$ denotes the movement and interpolation operation. This process is repeated $N$ times to construct the final point cloud $\mathcal{P}_N$. Through reprojection, $\mathcal{P}_N$ delivers high-quality, multi-view consistent images. The overall procedure for constructing $\mathcal{P}_N$ from $[\mathbf{I}_0, \mathbf{D}_0]$, $\mathbf{K}$.
\label{ssec:subhead2}

\vspace{-2pt} 
\subsection{Rendering with gaussian splatting}
After the point cloud is constructed, we train a 3D Gaussian Splatting model using the point cloud and the projected images. The centers of the Gaussian splats are initialized from the input point cloud, while the volume and position of each point are optimized under the supervision of the ground truth projected images.The loss function is constructed as a weighted combination of L1 loss and SSIM loss.  

Initializing with $\mathcal{P}_N$ accelerates network convergence and encourages the model to focus on reconstructing fine-grained details. For training, we use an additional set of $M$ images alongside the original $(N + 1)$ images employed in point cloud generation, as the initial set alone is insufficient for producing plausible results. These $M$ new images and their corresponding masks are generated by reprojecting $\mathcal{P}_N$ under a new camera sequence $\mathbf{P}_{N+1}, \dots, \mathbf{P}_{N+M}$:

\begin{equation}
\mathbf{I}_i, \mathbf{M}_i = f_{3d\rightarrow2d} \left( \mathcal{P}_N, \mathbf{K}, \mathbf{P}_i \right), \quad i = N+1, \dots, N+M.
\end{equation}

Note that we do not perform inpainting on $\mathbf{I}_i$ during the optimization of the Gaussian splats. Instead, when computing the loss function, we only consider the valid image regions where the mask value is 1. This prevents the model from learning artifacts in the reprojected images. Since each point is represented as a Gaussian distribution, missing pixels are naturally filled during rendering, and the resulting rasterized image becomes plausible after training.
\label{ssec:subhead4}

% \flbox{highlighter_red}{1em}\flbox{highlighter_orange}{1em}\flbox{highlighter_yellow}{1em}
\begin{table*}[t]
\centering
\caption{\textbf{WorldScore Benchmark Comparison.} Abbreviations: Ctrl=Controllability,Align=Alignment,Consist=Consistency,Photo=Photometric.The top three rankings from highest to lowest are marked with \showred  \showorange  \showyellow .}
\label{tab:worldscore}
\begin{tabular}{@{}l c ccc cccc@{}}
\toprule
\multirow{2}{*}{Models} & 
\multirow{2}{*}{Worldscore} & 
\multicolumn{3}{c}{Controllability} & 
\multicolumn{4}{c}{Quality} \\
\cmidrule(lr){3-5} \cmidrule(lr){6-9}
& & 
\makecell{Camera\\Ctrl} & \makecell{Object\\Ctrl} & \makecell{Content\\Align} & 
\makecell{3D\\Consist} & \makecell{Photo\\Consist} & \makecell{Style\\Consist} & \makecell{Subjective\\Qual} \\
\midrule
SceneScape\cite{19}      & 50.73 & 84.99 & \orangehl{47.44} & 28.64 & 76.54 & 62.88 & 21.85 & 32.75  \\ 
Text2Room\cite{20}      & 62.10 & \orangehl{94.01} & 38.93 & 50.79 & \yellowhl{88.71} & 88.36 & 37.23 & 36.69  \\ 
LucidDreamer\cite{21}    & \yellowhl{70.40} & 88.93 & \yellowhl{41.18} & \redhl{75.00} & \orangehl{90.37} & \orangehl{90.20} & 48.10 & \orangehl{58.99}  \\ 
WonderJourney\cite{22}   & 63.75 & 84.60 & 37.10 & 35.54 & 80.60 & 79.03 & \yellowhl{62.82} & \redhl{66.56}  \\ 
InvisibleStitch\cite{23} & 61.12 & \yellowhl{93.20} & 36.51 & 29.53 & 88.51 & \yellowhl{89.19} & 32.37 & \yellowhl{58.50}  \\ 
WonderWorld\cite{24}     & \orangehl{72.69} & 92.98 & \redhl{51.76} & \orangehl{71.25} & 86.87 & {85.56} & \orangehl{70.57} & 49.81  \\
\midrule 
Gen3d(ours)           & \redhl{75.05} & \redhl{99.77} & 41.00 & \yellowhl{66.73} & \redhl{91.79} & \redhl{91.11} & \redhl{77.01} & 57.92 \\ 
\bottomrule
\end{tabular}
\end{table*}

\begin{figure*}[!t]
\centering
\setlength{\tabcolsep}{6pt}
\renewcommand{\arraystretch}{2.5}

% 修正后的列格式定义
\begin{tabular}{
    >{\centering\arraybackslash}m{2cm}@{\hspace{6pt}}
    *{3}{>{\centering\arraybackslash}m{2.2cm}@{\hspace{10pt}}}
    @{\hspace{15pt}}|@{\hspace{15pt}}
    *{2}{>{\centering\arraybackslash}m{2.2cm}@{\hspace{10pt}}}
    >{\centering\arraybackslash}m{2cm}
}
% -------------------------- 表头 --------------------------
\multicolumn{1}{c@{\hspace{6pt}}}{} &
\multicolumn{3}{c@{\hspace{18pt}}|@{\hspace{15pt}}}{\textbf{Indoor}} &  % 修正了间距计算
\multicolumn{2}{c@{\hspace{12pt}}}{\textbf{Outdoor}} &
\multicolumn{1}{c}{} \\

% -------------------------- 第一行：Input --------------------------
\textbf{Input} & 
\includegraphics[width=2.5cm,valign=c]{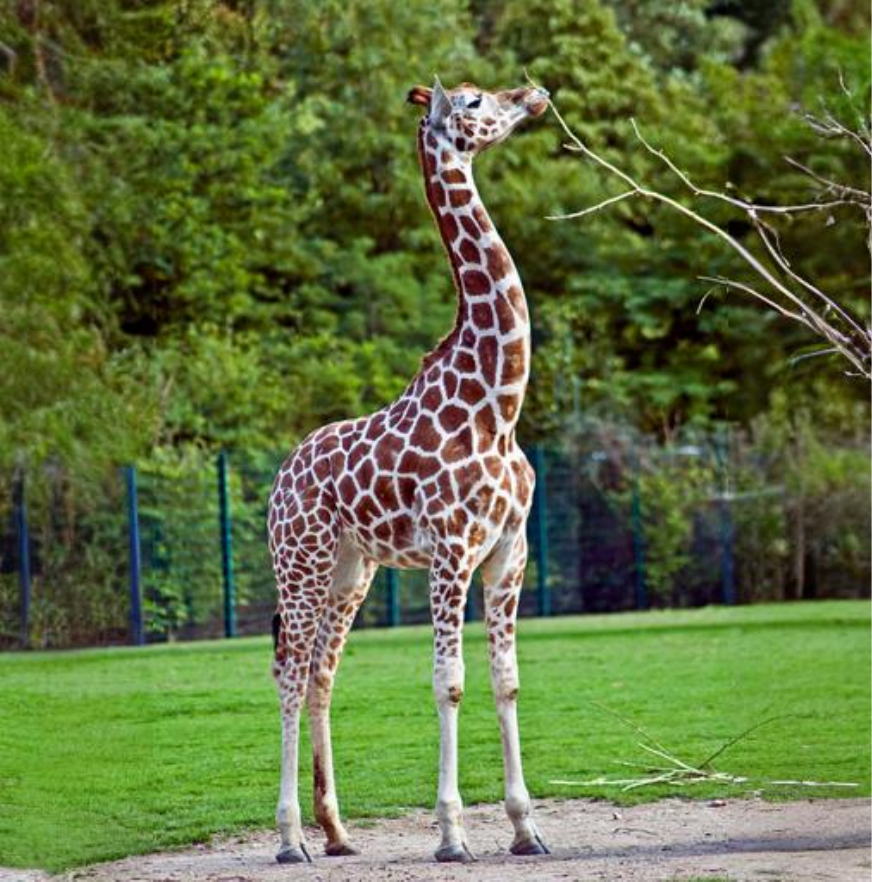} & 
\includegraphics[width=2.5cm,valign=c]{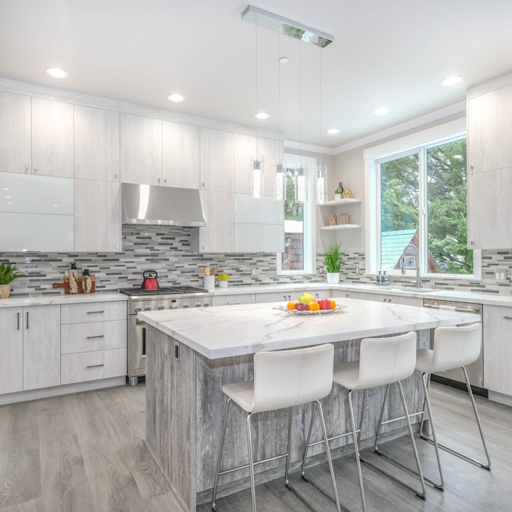} & 
\includegraphics[width=2.5cm,valign=c]{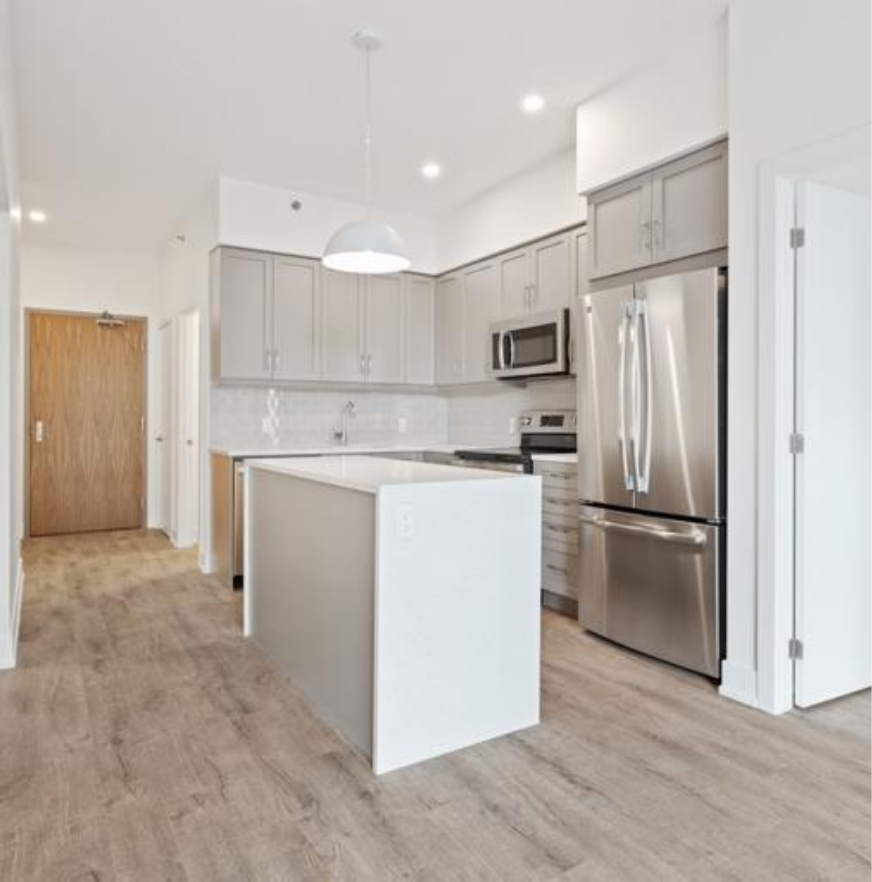} & 
\includegraphics[width=2.5cm,valign=c]{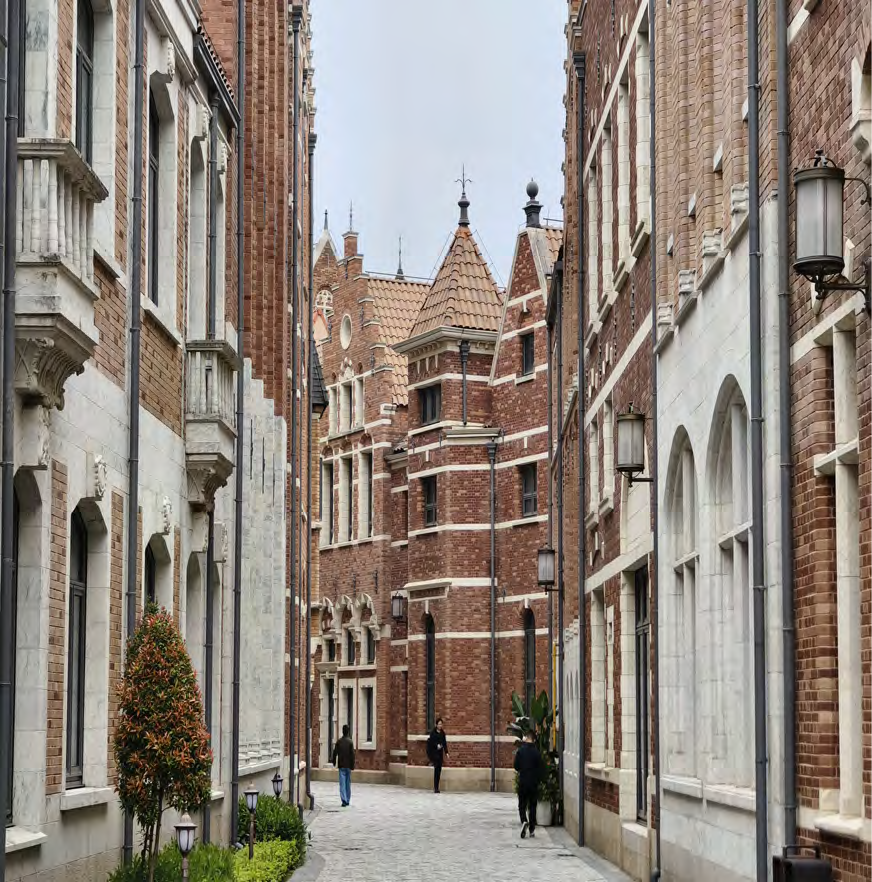} & 
\includegraphics[width=2.5cm,valign=c]{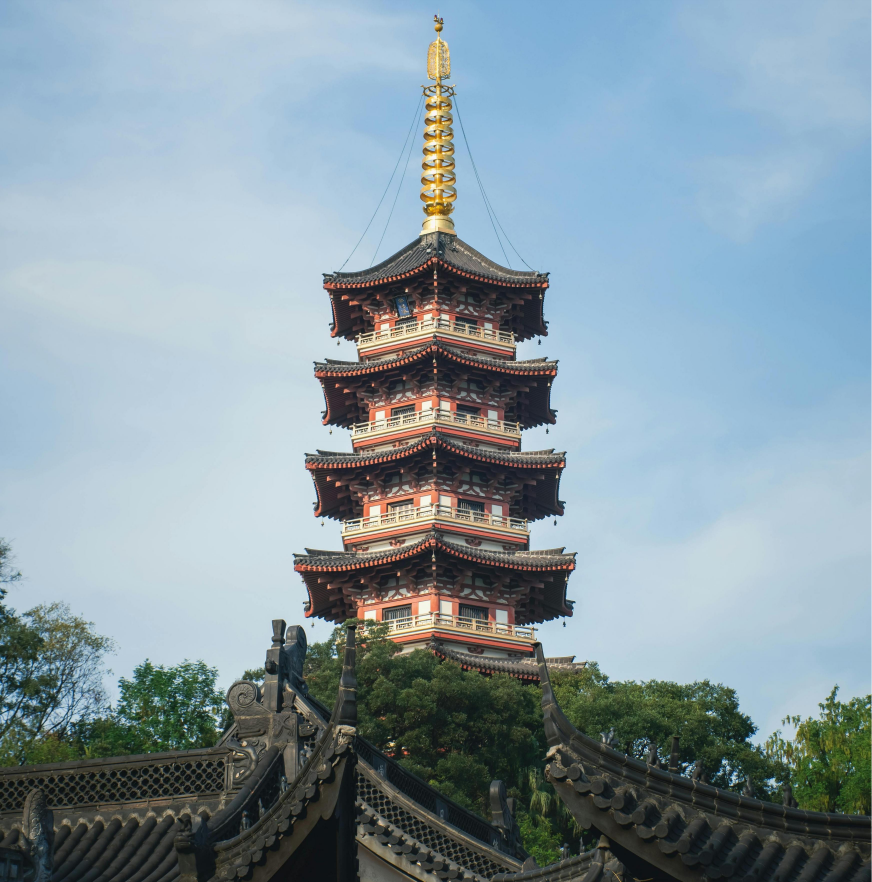} & 
\textbf{Input} \\
  
% -------------------------- 第二行：LucidDreamer / WonderWorld --------------------------
\textbf{LucidDreamer \newline \cite{21}} & 
\includegraphics[width=2.5cm,valign=c]{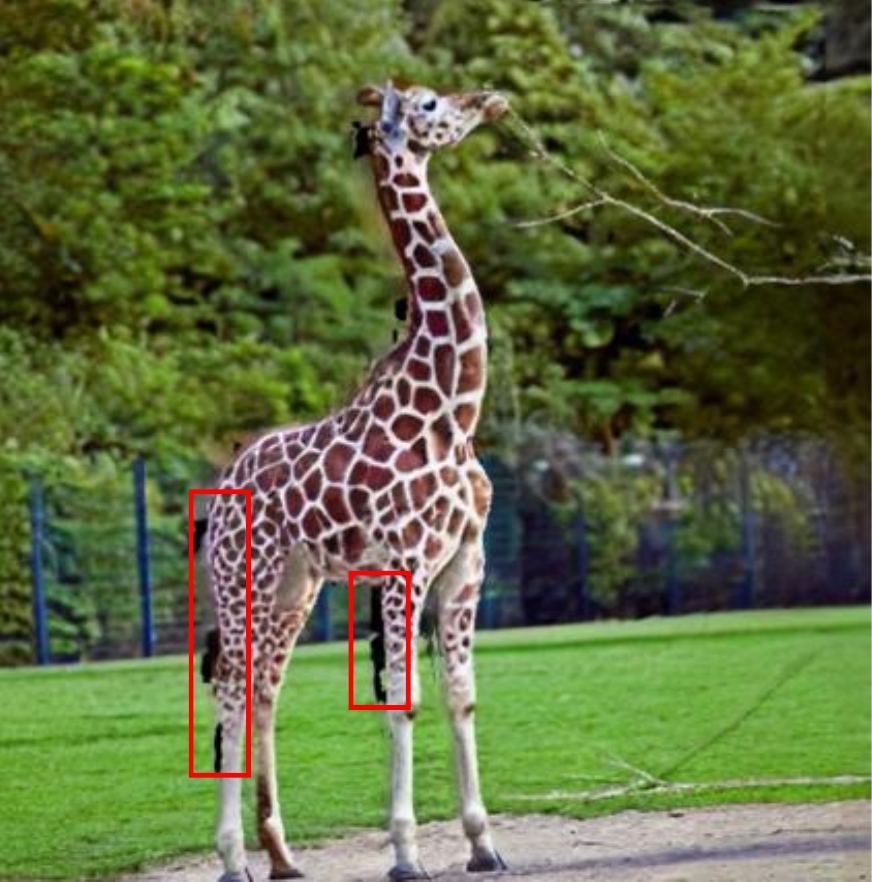} & 
\includegraphics[width=2.5cm,valign=c]{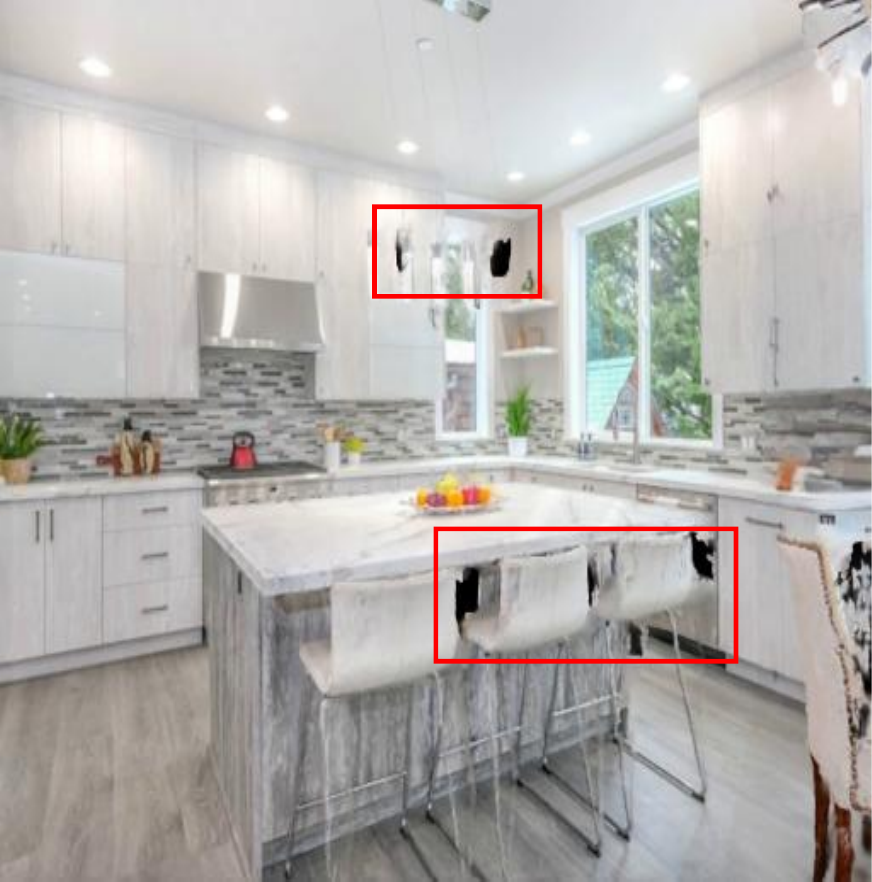} & 
\includegraphics[width=2.5cm,valign=c]{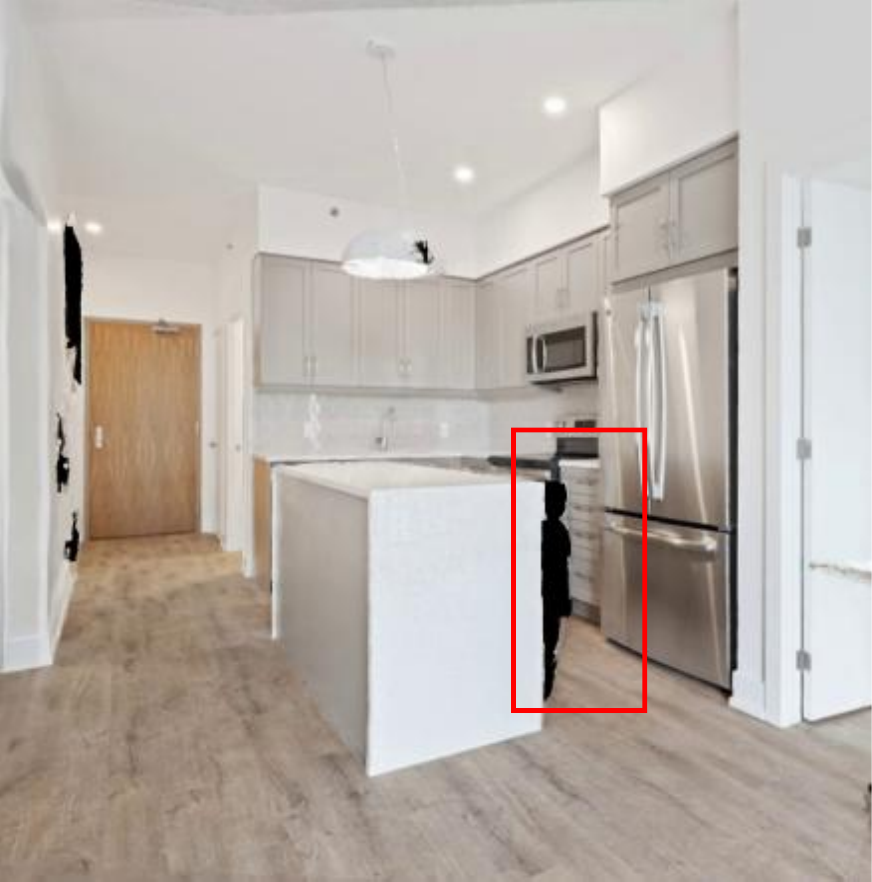} & 
\includegraphics[width=2.5cm,valign=c]{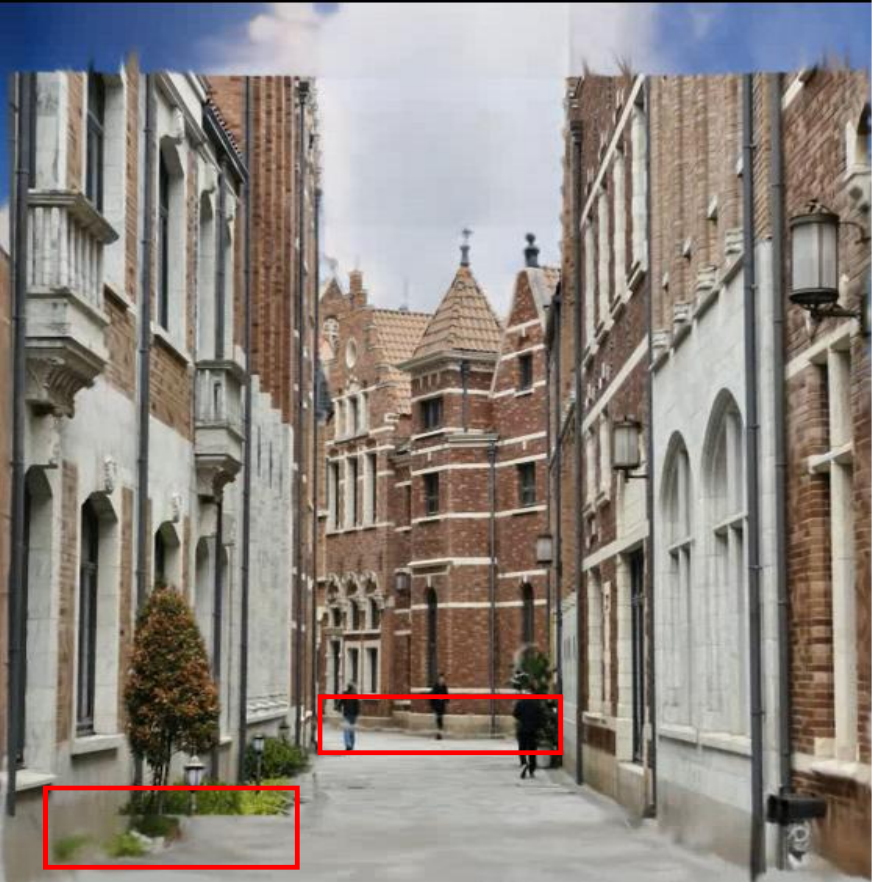} & 
\includegraphics[width=2.5cm,valign=c]{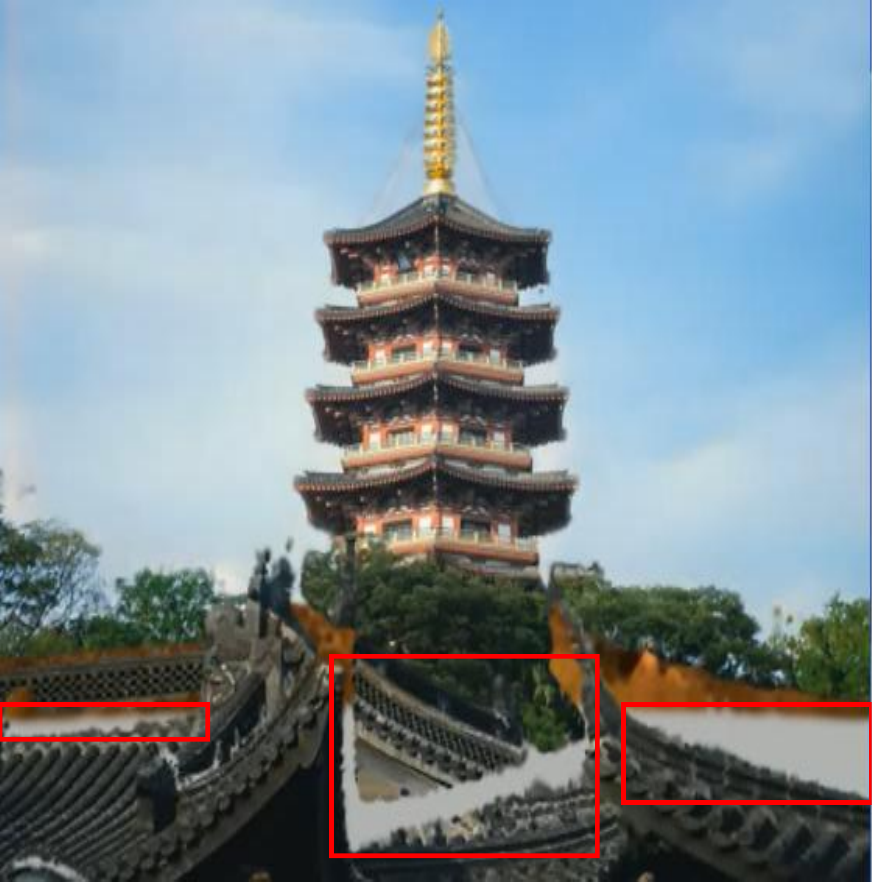} & 
\textbf{WonderWorld \newline \cite{24}} \\

% -------------------------- 第三行：Gen3d (Ours) --------------------------
\textbf{Gen3d (Ours)} & 
\includegraphics[width=2.5cm,valign=c]{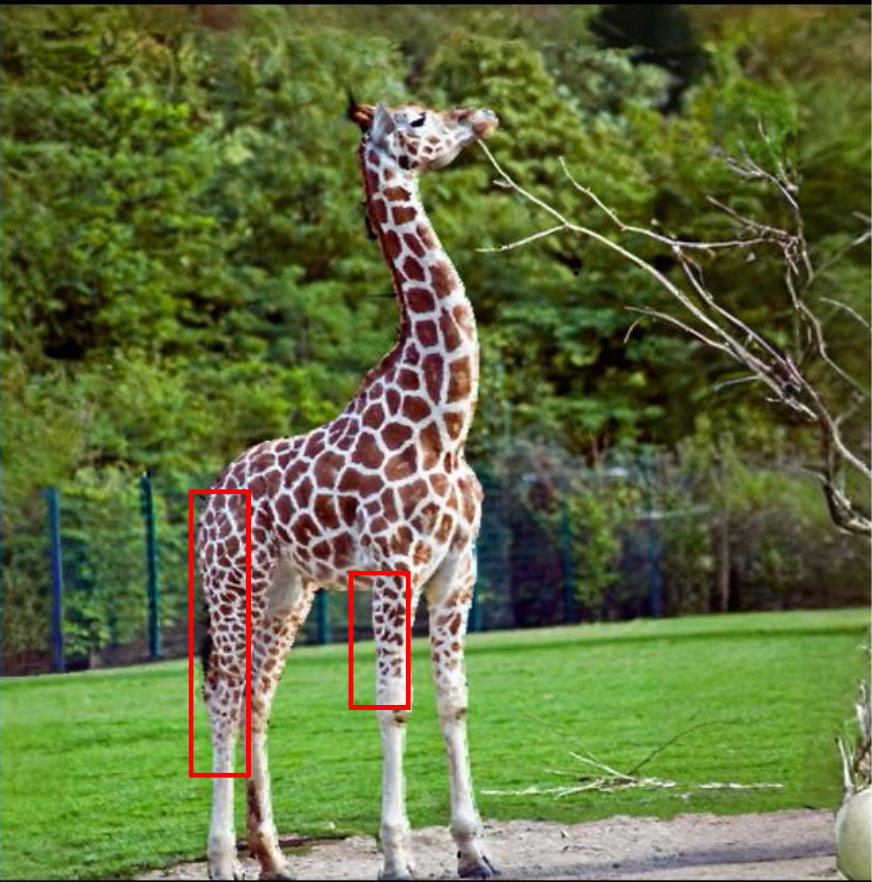} & 
\includegraphics[width=2.5cm,valign=c]{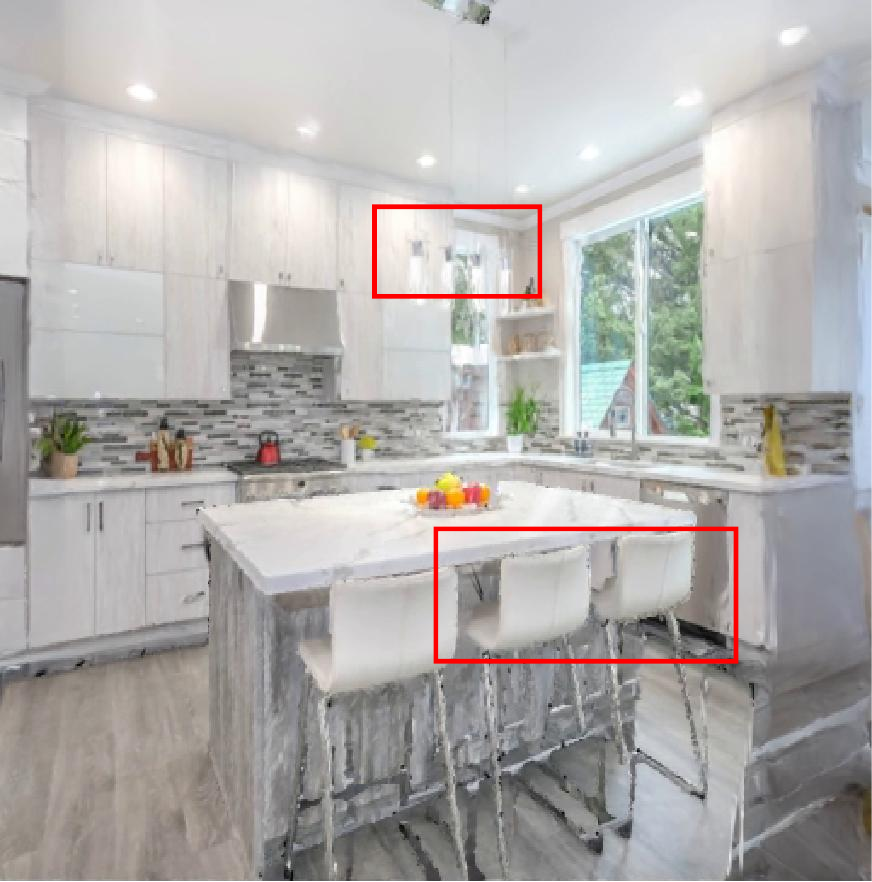} & 
\includegraphics[width=2.5cm,valign=c]{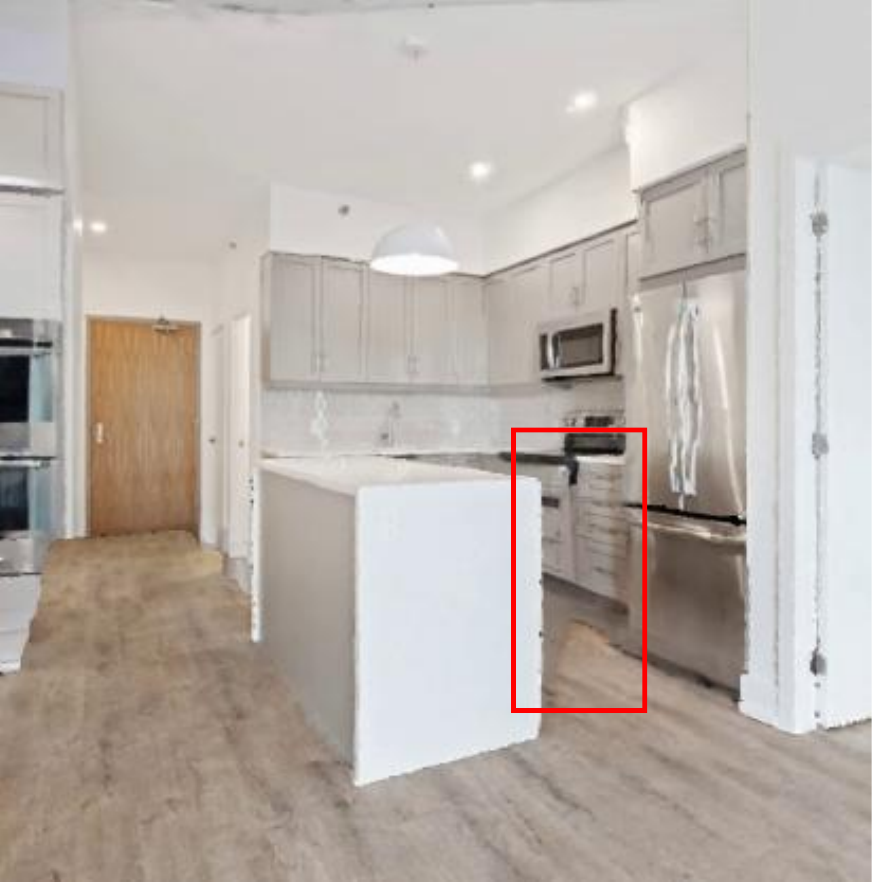} & 
\includegraphics[width=2.5cm,valign=c]{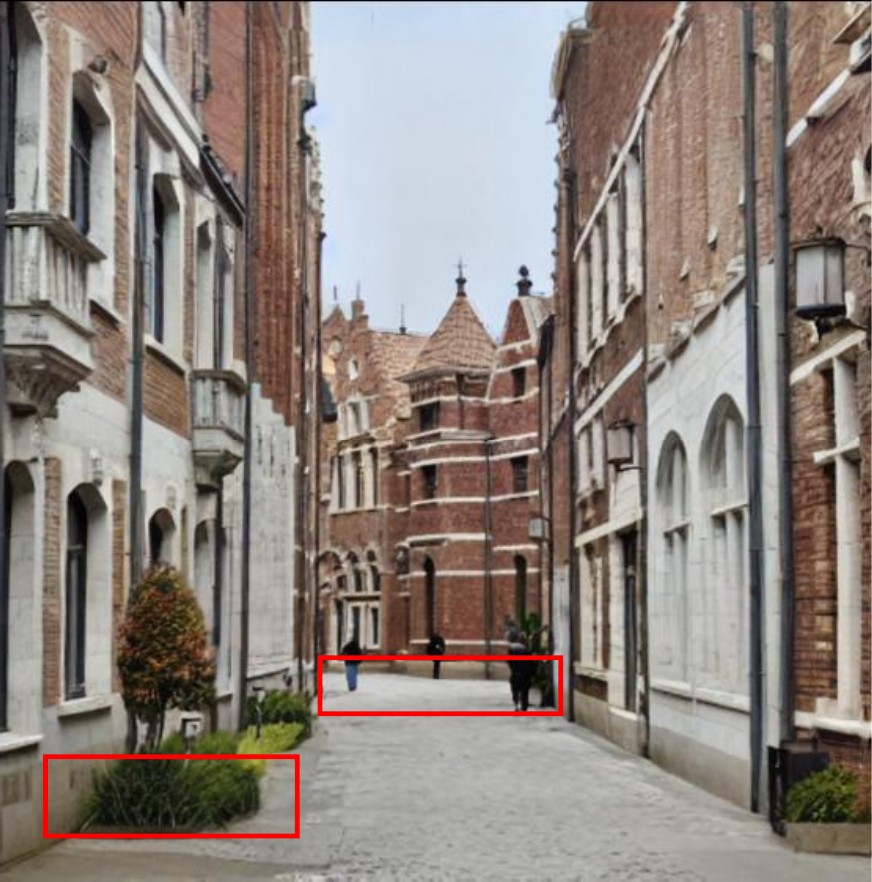} & 
\includegraphics[width=2.5cm,valign=c]{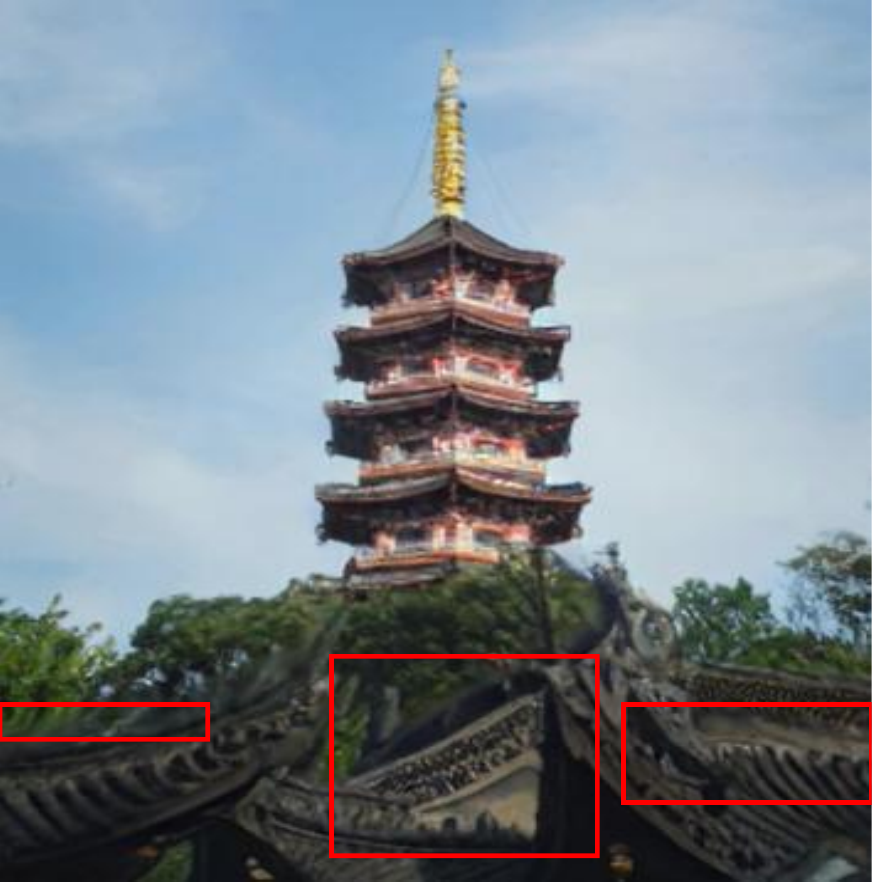} & 
\textbf{Gen3d (Ours)} \\
\end{tabular}

\caption{\textbf{Comparisons of multi-view generation results across different methods.}The images are sourced from  the COCO dataset, the WorldScore Benchmark, and web-sourced images.}
\label{fig:method_comparison_horizontal}
\end{figure*}

\section{Experiments}
\subsection{Implementation details.}
The modules we used to construct Gen3d can be either trained using manual design or brought from off-the-shelf models.We use pretrained large-scale off-the-shelf models to compose the
whole network to maximize the generalization capability
of the network. Specifically, we use the same text prompt input for the Stable Diffusion if the first image is generated from the text. If the input format is a RGB(D) image without text, we use LAVIS\cite{27}  to generate the caption according to the image and place it in the diffusion inpainting model to generate consistent content. For the camera trajectory that we use to construct the point cloud ($\{ \mathbf{P}_i \}_{i=0}^N$), we create several types of camera trajectory presets in advance, and different types of trajectories were used for different tasks.

\vspace{-2pt} 
\subsection{World Generation}
We test Gen3d on WorldScore \cite{25} static benchmark on world generation. WorldScore consists of 2,000 static test examples that span diverse worlds. The metrics evaluate the controllability and quality of video generation. Specifically, we use ``Camera Control'', ``Object Control'', and ``Content Alignment'' to judge how the model adhere to viewpoint instructions and text prompts. We use ``3D Consistency'', ``Photometric Consistency'', ``Style Consistency'', and ``Subjective Quality'' to evaluate the consistency and quality of generated content. Finally, an average score is presented to show the overall performance.  We compare six 3D generation methods in the existing benchmark. The scores are reported in Table\ref{tab:worldscore}.Gen3d achieves the highest score on this benchmark.The score shows that Gen3d has competitive performance on camera control and 3D consistency, compared with 3D Models.  Our Photometric consistency score is the highest among all methods, further demonstrating the visual quality of our generated videos. 

%Notably, our metric depth-based conditioning enables larger camera movements while maintaining superior consistency. 

To comprehensively evaluate our Gen3d model for multi-view 3D generation, we conducted qualitative comparisons with top-performing baselines in our benchmark: WonderWorld and LucidDreamer. These models were selected for their state-of-the-art performance.WonderWorld is optimized only for open-landscape outdoor scenes and fails at indoor generation. Thus, indoor comparisons only include LucidDreamer and Gen3d.

As shown in Fig. \ref{fig:method_comparison_horizontal}, LucidDreamer has two critical flaws under large camera movements: noticeable geometric holes in complex occlusions, and inconsistent generation quality across views. In contrast, Gen3d via our layered strategy  decomposes scenes into layers and optimizes their consistency independently, effectively synthesizing occluded regions and maintaining texture and shape uniformity. For outdoor scenes, WonderWorld, despite its strengths in clear-sky outdoor environments, still produces obvious floating artifacts during camera motion owing to the use of semantic hard-coding for reconstruction.Gen3d, however, maintains stable, artifact-free generation and accurate geometry in these challenging outdoor scenarios.

\section{Conclusion}
% This study addresses the limitation of traditional neural 3D reconstruction—reliance on dense multi-view captures—by proposing Gen3d, an end-to-end pipeline for high-quality, wide-scope 3D scene generation from a single input (text, RGB, or RGBD).

% Gen3d integrates three core technologies: Stable Diffusion for photorealistic 2D synthesis, 3D Gaussian Splatting (3DGS) for efficient 3D representation, and point cloud geometric guidance for cross-view consistency. It proceeds in three key steps: foreground-background segmentation with initial point cloud generation, iterative point cloud expansion via camera trajectory simulation and diffusion-based novel view synthesis, and 3DGS optimization to eliminate holes and enable real-time rendering.

% % Experiments on  WorldScore show Gen3d outperforms baselines  with superior image quality, pose control, and domain generalization. It also supports multi-modal input fusion and dynamic input adjustment.

% Future work will focus on accelerating point cloud expansion and extending Gen3d to dynamic 3D scene generation.

This study addresses the limitation of traditional neural 3D reconstruction—reliance on dense multi-view captures—by proposing Gen3d, a novel pipeline for high-quality, wide-scope 3D scene
generation from a single input.

Gen3d includes: Stable Diffusion for photorealistic 2D synthesis, 3DGS for efficient 3D representation, and point cloud geometric guidance for
cross-view consistency.It proceeds in steps: segmentation with initial point cloud generation, iterative point cloud expansion,
diffusion-based novel view synthesis,3DGS optimization to
eliminate holes and enable real-time rendering.

Future work will focus on accelerating point cloud expansion
and extending Gen3d to dynamic 3D scene generation.

\bibliographystyle{IEEEbib}
\bibliography{references}
\end{document}